**A bilingual approach to specialised adjectives through word embeddings in the karstology domain**

(The paper will be published as part of TOTH 2020 proceedings http://toth.condillac.org/proceedings)

Larisa Grčić Simeunović[1], Matej Martinc[2], Špela Vintar[3]

University of Zadar[1], Jožef Stefan Institute [2], University of Ljubljana[3]

M. Pavlinovića 1, HR-23000 Zadar[1], Jamova cesta 39, SI-1000 Ljubljana[2], Aškerčeva 2, SI-1000 Ljubljana[3]

lgrcic@unizd.hr, matej.martinc@ijs.si, spela.vintar@ff.uni-lj.si

**Abstract**

We present an experiment in extracting adjectives which express a specific semantic relation using word embeddings. The results of the experiment are then thoroughly analysed and categorised into groups of adjectives exhibiting formal or semantic similarity. The experiment and analysis are performed for English and Croatian in the domain of karstology using data sets and methods developed in the TermFrame project. The main original contributions of the article are twofold: firstly, proposing a new and promising method of extracting semantically related words relevant for terminology, and secondly, providing a detailed evaluation of the output so that we gain a better understanding of the domain-specific semantic structures on the one hand and the types of similarities extracted by word embeddings on the other.

## 1      Introduction

In this paper we explore the bilingual comparative approach to adjectives through word embeddings in the domain of karstology. Because specialized adjectives have not received much attention in the field of terminology so far we aim at modelling the semantic relations expressed by adjectives within multiword terms and discovering their shared properties through word embeddings. Our starting hypothesis is that adjectival semantic relations are an important part of conceptual representations as they establish links between terms and their attributes inside conceptual frame. For this reason, we aim to reveal adjectives expressing exemplar characteristics in order to establish attribute-value pairs for different conceptual categories in karst domain.

By leveraging word embeddings and sets of seed adjectives expressing specific semantic relations we aim to extract additional adjectives that express the same relation and rate the degree of semantic similarity between adjectives in the two languages. Furthermore, we perform a detailed analysis of the extracted candidates in terms of their semantic and morphosyntactic properties in order to identify corresponding clusters between adjectives, but also to better understand the errors produced by our prediction algorithm.

Karstology is an interdisciplinary domain which studies karst, a type of landscape developing on soluble rocks such as limestone or gypsum. The most distinctive features of karst regions include caves, various types of relief depressions, submerged rivers, springs, ponors and sinkholes. The TermFrame project models the karstology domain using the frame-based approach (Faber, 2012; Faber, 2015). To explore typical conceptual frames in karstology we developed a domain-specific concept hierarchy of semantic categories, and each category can be described by a set of relations which reveal its specific features. In addition to manually identified categories and relations we employ a number of advanced text mining techniques to extract structured domain knowledge from our corpora (Miljković et al., 2019; Pollak et al., 2019; Vintar et al., 2020, Pollak et al. 2020). The final result of the project will be a multilingual visual knowledge base for karstology.

The experiment and analysis presented in this paper seek to highlight the synergies and parallels between the frame-based and cognition-inspired view of specialised language on the one hand, and on the other hand word embeddings as a mathematical device to map meaning into a multidimensional space. It is no coincidence that we perceive cognition as a dynamic and inherently spatial operation, and recent advances in deep learning for NLP prove that conceptual similarities are indeed reflected in the spatial proximity or closeness of word embeddings. By performing a detailed analysis of the adjectives extracted through embeddings we aim to shed light on the different nuances of semantic similarity as computed via deep learning methods.

The paper is structured as follows: In Section 2 we present related research, Section 3 describes the methods used for the extraction experiment and manual analysis, and Section 4 presents the analysis of adjectival clusters for each semantic relation and language respectively. We conclude with a discussion and final observations.

## 2   Related work

The representation of relations between concepts has already been presented in different terminological works which aim to illustrate the dynamics of cognition. From the point of view of Frame-Based Terminology (Faber et al. 2007; Faber and Leon Araúz 2016; Gil-Berrozpe et al. 2019; Cabezas-García and Leon Araúz 2018) knowledge structures are organized as frames on the basis of elements and entities which share similar contexts and situations. The relations between them allow us to construct meaningful schemes or mental representations of segments that belong to a particular specialized domain.

Even though previous work on exemplar models does not include explicit representation of attributes and their values, Barsalou (1992: 25) considers that recognising the values of the same attribute relies on the "embedded level of exemplar processing for categorizing characteristics as values of attribute categories ".

According to Petersen (2015), "[t]he attributes in a concept frame are the general properties or dimensions by which the respective concept is described"[1]. Their values express specifications that are important for creating concept frames or models for the representation of concepts.  Such frames can be used as templates to guide the formulation of definitions (Duran Muñoz, 2016), or they can be applied to predict the semantic category of an attribute in a multi-word term, as our study illustrates.

---

[1] The attributes are defined by Barsalou (1992:30) as „a concept that describes an aspect of at least some category members. For example, color describes an aspect of birds. (…) A concept is an attribute only when it describes an aspect of a larger whole. When people consider color in isolation (e. g., thinking about their favorite color), it is not an attribute but is simply a concept".

According to Baroni et al. (2014), context predicting models are a promising way for performing a number of experiments on different semantic similarity tasks and datasets such as semantic relatedness, synonym detection, concept categorisation, selectional preferences and analogy.

Diaz et al. (2016) and Pollak et al. (2019) who conducted previous research on set expansion tasks showed that embeddings can be successfully employed for query expansion on domain specific texts.

## 3      Methodology

For the purposes of our research, we use the English and Croatian part of the TermFrame corpus which contains relevant contemporary works on karstology and is representative in terms of the domain and text types included. It comprises scientific texts (scientific papers, books, articles, doctoral and master's theses, glossaries and dictionaries) from the field of karstology, which in itself is an interdisciplinary domain partly overlapping with surface and subsurface geomorphology, geology, hydrology and other fields. Table 1 gives basic information about the corpus.

|           | English   | Croatian  |
|-----------|-----------|-----------|
| Tokens    | 2,721,042 | 1,229,368 |
| Words     | 2,195,982 | 969,735   |
| Sentences | 97,187    | 53,017    |
| Documents | 57        | 43        |

Table 1: Corpus information

In our previous research (Vintar et al., 2020) we proposed a method to extract expressions pertaining to a specific semantic relation from a comparable English and Croatian corpus by providing a limited number of seed words for each language and relation, then using intersections of word embeddings to identify words belonging to same relation class. An evaluation of the extracted candidates showed high variability in precision between relations and languages, ranging from 0.28 (FUNCTION, Croatian) to 0.80 (COMPOSITION, Croatian).

In this study we continue our analysis by exploring contexts where adjective-noun phrases are used to express semantic relations, by grouping the extracted adjectives into clusters according to their semantic and morphosyntactic properties, and by analysing erroneously extracted candidates. First, we classify the adjectives according to their semantic relation guided by the conceptual frame. The semantic relations of the adjectives are determined according to their dominant meaning in the domain of karst. For the purposes of this analysis adjectives are assigned to one of the 5 semantic relations: LOCATION (*underground* cave), CAUSE (*fluvial* sediment), FORM (*vertical* shaft), COMPOSITION (*gypsum* karst) and FUNCTION (*soluble* rock).

Nevertheless, it is important to bear in mind that this classification is not unambiguous as adjectives can take different meanings depending on the nouns they modify. For example, in a phrase like *korozijski proces* the value *korozijski* is assigned to the relation FUNCTION, while in a syntagm *korozijska ponikva* the value of the adjective is CAUSE.

Secondly, we investigate the extraction of adjectives from word embeddings trained on English and Croatian specialised corpora. Specifically, the task is to find adjectives that qualify their respective headwords along 5 fixed semantic dimensions we refer to as relations, using a set of seed adjectives for each relation and language. We trained FastText embeddings (Bojanowski et al., 2017) on the English and Croatian part of the TermFrame corpus respectively, using the skip-gram model with the embedding dimension of 100. For each seed adjective expressing a specific semantic relation, we extract a set of 100 closest words according to the cosine distance. Then, we calculate intersections between these sets of closest words, for all combinations of seed adjectives and subset sizes 2 – 10 (see Vintar et al., 2020).

|   | location | | function | | form | | composition | | cause | |
|---|---|---|---|---|---|---|---|---|---|---|
|   | en | cr | en | cr | en | cr | en | cr | en | cr |
| N | 357 | 228 | 147 | 152 | 164 | 152 | 293 | 244 | 183 | 181 |
| C | 118 | 88 | 68 | 43 | 108 | 97 | 184 | 197 | 88 | 132 |
| P | 0.33 | 0.39 | 0.46 | 0.28 | 0.66 | 0.64 | 0.63 | 0.80 | 0.48 | 0.73 |

Table 2: Precision per semantic relation and language (N = number of extracted words, C = correct, P = precision)

The results show the overall lowest and highest precisions in both languages as well as large differences between individual semantic relations. The prediction of semantic class membership is confirmed even for expressions with very low frequency.

In Section 4 we present the results of a manual analysis of adjectives extracted through the embeddings intersection method. In particular, we look for clusters of morphosyntactically, semantically or derivationally related words and look for patterns of similarity across the two languages. To perform our analysis, we introduce suffix and prefix-based clusters as well as derivational cluster. The proposed method helps improve the interpretability of the embedding results.

## 4    Analysing patterns of similarity

In the following subsections we attempt to categorise the extracted adjectives into groups or clusters, whereby we consider the seed adjectives provided for each semantic relation prior to the extraction task. We present results for English and Croatian respectively, then a comparison is made between findings.

4.1     CAUSE

4.1.1 English

Seed adjectives: *allogenic*, *anthropogenic, fluvial, alluvial, erosional, solutional, periglacial, tectonic, volcanic, lacustrine, aeolian*

a) Suffix-based clusters corresponding seed words:

*allogenic, anthropogenic:* -**genic**: epigenic, geogenic, cryogenic, autogenic, orogenic, biogenic, pathogenic, hypogenic, glacigenic, monogenic, rheogenic, speleogenic, radiogenic, guanogenic

*fluvial, alluvial:* -**luvial**: eluvial, colluvial, pluvial, deluvial,

*periglacial:* -**glacial**: preglacial, subglacial, fluvioglacial, englacial, proglacial, supraglacial, postglacial, paraglacial, pleniglacial, glaciofluvial,

-**al** *(erosional, solutional)*: disolutional, denudational, compressional, tensional, gravitational, lagoonal, formational, corrosional, depositional, torrential, detrital, deglacial, abrasional, suffosional, evolutional, dissolutional, subaerial

b) Suffix-based clusters not corresponding seed words:

-**ous:** terrigenous, autochthonous, calcareous, argillaceous, igneous

-**clastic**: thermoclastic, volcanoclastic, bioclastic, pyroclastic, clastic, siliclastic, siliciclastic

-**karstic**: glaciokarstic, fluviokarstic,

-**genetic:** paragenetic,

**No group**: lacustrine

4.1.2 Croatian

Seed adjectives: *alogen, antropogen, fluvijalan, erozijski, aluvijalan, vulkanski, lakustrijski, eolski, periglacijalni, tektonski*

a) Suffix-based clusters corresponding seed words:

*alogen, antropogen:* **-gen**: egzogen, kemogen, zoogen, biogen, kriogen, epigenijski, orogenski;

*fluvijalan, aluvijalan:* **-luvijalan** and **-fluvijalan**: iluvijalan, proluvijalan, delovijalan, diluvijalan, koluvijalan, glaciofluvijalan, postfluvijalan;

*periglacijalan:* **(-glacijalan)**: glacijalan, proglacijalan, interglacijalan, postglacijalan, fluvioglacijalan;

-**ski** *(erozijski, eolski)*: abrazijski, mikroerozijski, disolucijski, denudacijski, fluviodenudacijski, derazijski, destrukcijski, dislokacijski, soliflukcijski, subdukcijski, inundacijski, marinski, međuzrnski, siparski, eforacijski, amfibolski, supsidencijski, egzarazijski, padinski, mindelski, evolucijski, monoklinski, piraterijski, magmatski, evorzijski, melioracijski, kriofrakcijski, translacijski, poligonski, riski, oligocenski, intrabazenski, plutonski, drobinski, akumulacijski, superpozicijski, litotamnijski, submarinski, regresijski, alveolinski, osulinski

**-an, -ni**: erozivan, abrazivan, piroklastični, naplavni, terasan, stadijalan, riječni, šljunčani, žilni, bazalni, bazaltni, pretaložen, denudiran, hipoabisalan, nataložen, bujičan, naplavljen, bujičav, pleistocenalan;

b) Suffix-based clusters not corresponding seed words:

-**čan** : klastičan, vulkanoklastičan, piroklastičan;

**-genetski** : biogenetski, klimamorfogenetski, tektogenetski, epigenetski, poligenetski, epirogenetski;

Observations:

Starting from the seed words in two languages we can observe high correspondence in results for this relation. Clusters around the same suffixes are formed in both languages:

*-genic (allogenic, anthropogenic)* (14) and *-gen (alogeni, antropogeni)* (7),

*-luvial (fluvial, alluvial)* (4) and *-luvijalni (fluvijalni, aluvijalni)* (8)

*-glacial (periglacial)* (10) *and -glacijalni (periglacijalni)* (5)

The remainder of the results also exhibit productive suffixes that can be specific for this semantic relation: the suffixes **-al, -ous,** -**clastic**, -**karstic, -genetic** in English and the suffixes **-ski, -an, -ni,** -**čan, -genetski** in Croatian.

This confirms that embedding methodology successfully retrieves adjectives from the same semantic relation.

## 4.2 COMPOSITION

4.2.1 English

Seed adjectives: *carbonate, limestone, dolomitic, sedimentary, sulphate, calcareous, carboniferous, silicate, sulphuric, diagenetic, siliceous, clay*

a) Suffix-based clusters corresponding seed words:

*carbonate:* metacarbonate, noncarbonate, bicarbonate;

*limestone:* **-stone:** rimestone, grainstone, dolostone, framestone;

*dolomitic, sulfuric, diagenetic:* **-ic:** evaporitic, quartzitic, conglomeratic, loessic, calcitic, sulphuric, silicic, pelitic, andesitic, carbonatic, carbonic, bioclastic, granitic, basaltic, magmatic, metallic, elastic, aphanitic, ophiolitic;

*calcareous, carboniferous, siliceous:* **-ous:** gypsiferous, argillaceous, igneous, amorphous, herbaceous, carboniferous, siliceous, carbonaceous, fossiliferous, sulfurous, tufaceous;

*silicate, sulfate:* **-ate**: vanadate, cipitate, dehydrate;

b) Suffix-based clusters not corresponding seed words:

**-ite:** siderite, anhydrite, hexahydrite, bassanite, phyllite, kaolinite, tyuyamunite, biomicrite, biopelmicrite, kimberlite, phosphorite, micrite, halloysite, serpentinite, dickite, alterite, barite, laterite;

-**mineral**: thermomineral, monominerallic;

-**grained:** finegrained;

No group: shale, schist, sandy, oxidase, paleokarstic, sulfide, silty, zechstein, claypan, foraminifer, manganese, haematite, chalky, nonsoluble, interclas, silicify, phyllito, monocrystalline, oxide,

4.2.2 Croatian

Seed adjectives: *karbonatan, vapnenački, dolomitski, sedimentan, kalcitan, karbonski, sulfatan, glinovit, sedren, stijenski*

a) Suffix-based clusters corresponding seed words:

*karbonatan:* nekarbonatan, hidrokarbonatan;

*kalcitan, sedimentan, sulfatan:* **-an**: aragonitan, detritičan, pješčan, kristaličan, rudistan, silikatan, evaporitan, gipsan, flišan;

*dolomitski, karbonski, stijenski:* **-ski:** alveolinski, amfibolski, drobinski, foraminiferski, morenski;

*vapnenački*: **-čki**: škriljevački;

*glinovit*: **-it:** laporovit, muljevit, pjeskovit, šljunkovit*;*

b) Suffix-based clusters not corresponding seed words:

*zrnat*: krupnozrnat, međuzrnski, sitnozrnat, zrnat/ zrnast;

Observations:

The embedding candidates in English can be organised according to productive suffixes like **-ic, -ous, -ate, -clastic, -ite,** -**mineral, -grained.**

In Croatian suffixes **-an/-ni, -ski, -čki, -it, -zrnat** are productive for the semantic relation composition.

**4.3 FORM**

4.3.1 English

Seed adjectives: *polygonal, vertical, dendritic, shallow, enclosed, elongated, flat, steep, cavernicolous, detrital*

a) Suffix-based clusters corresponding seed words:

*polygonal, vertical*: **-al**: elliptical, elliptic, cylindrical, subparallel, subhorizontal, subvertical, centripetal, symmetrical, subhorizontal, sinusoidal, asymmetrical, orthogonal, rectilinear, concave, angular, rectangular;

*cavernicolous:* **-ous**: sinuous;

No group: staircase, meander, singular, elbow, cylinder, labyrinthine, sharply, pinnacle, steeply,

4.3.2 Croatian

Seed adjectives: *vertikalan, ravnocrtan, strm, kavernozan, horizontalan, mrežast, longitudinalan, kružan, razgranat, ulegnut, uravnjen*

a) Suffix-based clusters corresponding seed words:

*vertikalan, ravnocrtan, horizontalan, longitudinalan:* **-an**: konveksan, tangencijalan, trodimenzionalan, subhorizontalan, simetričan, asimetričan, paralelan, tlocrtan, konkavan, ustrmljen, ovalan, nepravilan, polukružan, vodoravan, cilindričan, centrifugalan, centripetalan, konusan, dijagonalan, lateralan, longitudinalan, horizontalan, radijalan, konvergentan, etažan, urušan;

*mrežast:* **-ast**: ravničast, stepeničast, zvjezdast, žljebast, grozdast, prstenast, rešetkast, laktast, klifast, bunarast, dolinast, ponikvast, sigast, stepeničast, terasast;

b) Derivational cluster

Adjectives like *kavernozan, prevjesan, sinklinalan, monoklinalan, abisalan, fleksuran* are derived from the terms denoting karst forms:

kavernozan<kaverna, prevjesan<prevjes, sinklinalan<sinklinala, monoklinalan<monoklinala, abisalan<abis, fleksuran<fleksura;

No group: valovit, meandrirajući.

Observations: The results in this semantic relation exhibit clusters around seed adjectives *vertical, polygonal* in English and *vertikalan, ravnocrtan, horizontalan, longitudinalan* in Croatian. These adjectives refer to different shapes that are frequent in karst relief but can also refer to different other entities in nature.

Results in Croatian demonstrate a group of relational adjectives derived from karst forms: kavernozan<kaverna, prevjesan<prevjes, sinklinalan<sinklinala, monoklinalan<monoklinala, abisalan<abis, fleksuran<fleksura.

The suffixe **-ast** is also productive for expressing karst shapes in Croatian.

We can observe that the embeddings missed to propose adjectives that refer to karst forms. For. ex. *bunarast, ponikvast, terasast, krovinski, sigasti, zaravnjen, dolinast, škrapski*

### 4.4 FUNCTION

4.4.1 English

Seed adjectives: *impermeable, permeable, solutional, hydrothermal, speleological, geological, soluble, porous, depositional, regressive, undersaturated*

a) Suffix-based clusters corresponding seed words:

*soluble:* **-ble**: insoluble, impenetrable, rechargeable;

*porous:* **-ous**: anhydrous, impervious;

*speleological, geological:* **-logical**: paleontological, speleological, seismological, speleogical, petrological, biospeleological, climatological, vulcanospeleological, sedimentological, karstological;

Results ending in **-logical** but not connected to the karst domain: immunological, archeological, meteorological, histological, palynological, psychological, methodological, mythological,

**No group**: unkarstify, evaporitic, aquifer, dissociate, unsaturate, unconformable, lithoclast, friable, diffuse;

4.4.2 Croatian

Seed adjectives: *nepropustan, propustan, speleološki, geološki, topiv, porozan, taložan, urušan*

a) Suffix-based clusters corresponding seed words:

*topiv:* **-iv:** netopiv, vodotopiv, vododrživ, vododržljiv, propustljiv;

*porozan, taložan, urušan, nepropustan, propustan:* **-an:** vodonepropustan, vodopropusan, vodoprohodan, nepropusan, polupropusan, laminaran, difuzan, procjedan;

*speleološki, geološki:* **-ški:** speleomorfološki, geomofološki, etnološki, geoekološki, arheološki, aerološki, fiziološki, geoekološki, geokronološki, biološki, paleontološki,

No group: kemogen

Observations:

Seed adjectives *soluble, porous* in English and *topiv, porozan* in Croatian create suffix-based clusters in **-ble**, **-ous** in English and in -**iv**, **-an** in Croatian. While the results in English missed to propose the adjective *permeable*, results in Croatian demonstrate productive cluster around the seed adjectives *propustan, nepropustan.*

The suffixes **-logical, -loški** are productive in both languages since they refer to general concepts such as speleology, karstology, paleontology, seismology, slimatology, sedimentology.

### 4.5. LOCATION
#### 4.5.1 English

**Seed adjectives**: *coastal, littoral, sublittoral, submarine, oceanic, subsurface, subterranean, subterraneous, subaerial, underground, aquatic, subaqueous, internal, subglacial, phreatic, epiphreatic, vadose*

a) Suffix-based clusters corresponding seed words:

*coastal, littoral:* **-al**: sublittoral, paralittoral, abyssal, intracontinental, continental, peripheral;

*phreatic, epiphreatic:* **-ic:** bathyphreatic;

b) Suffix-based clusters not corresponding seed words:

**-ic:** meteoric, aquatic, semiaquatic, atlantic, pacific, anastomotic;

**-al**: interfluvial, terrestrial, superficial;

**-most**: uppermost, lowermost;

**-flow**: underflow, downflow;

**-haline**: polyhaline, sakhalin, euhaline, anchihaline, mixohaline;

**-shore**: seashore, offshore;

c) Prefix-based clusters corresponding seed words:

*subterraneous, subaqueous, subaerial, submarine, subsurface:* **sub-:** subterranean, subvertical, subtidal, subhorizontal, subzone;

d) Prefix-based clusters not corresponding seed words:

**sea-**: seafloor, seawater;

**hypo**-: hyporheic, hypokarst, hyporheal, hyporheo;

**hyper-:** hyperkarst;

**Other:** cavernicole, microcavernicole, cavernicolous, anastomose, aquaticus, shallowwater, saline, waterline, shoreline, ocean, subjacent, branchwork, mesovoid, lacustrine, intrastratal, streamway, surfacedwelling, crevicular, supratidal, interstice, marine, nonmarine,

### 4.5.2 Croatian

**Seed adjectives**: *obalan, litoralan, priobalan, podmorski, oceanski, podzeman, vadozan, podvodan, dolinski, špiljski, freatski, epifreatski*

a) Suffix-based clusters corresponding seed words:

*obalan, litoralan, priobalan, vadozan, podvodan:* **-an:** maritiman, bazalan, lateralan, inversan, otočan, dugovalan, saturiran;

*podmorski, oceanski, dolinski, špiljski, freatski, epifreatski:* **-ski:** zonski, primorski, dubokomorski, prekomorski, plitkomorski, obalski, priobalski, kontinentski, litoralizacijski, piedmontski, dubinski, sifonski, jamski, zavalski, ekvatorski, ponorski, kanalski, vršinski,;

Many of the adjectives ending in **-ski** are derived from karst forms: jamski<jama, zavalski<zavala, ekvatorski<ekvator, ponorski<ponor, kanalski< kanal,

b) Prefix-based clusters not corresponding seed words:

**sub-**: submarinski, subfreatski, subaerski;

**Observations:** Results show that this relation is not as precise as the others because embedding candidates cannot be interpreted univocally as locations and great number of mistakes lead us to consider the multidimensionality of the adjectives. In the next section we explain this dilemma in more detail.

### 4.6 Adjectives assigned to the wrong relation

As the results above show, the method based on word embeddings can be of broad use in pattern recognition. However it is also important to bear in mind the mistakes that appear along. Most of the mistakes can be found within the semantic relations of LOCATION, FUNCTION and COMPOSITION.

4.6.1 LOCATION

The adjective candidates that evidently do not determine location are the following:

- **genetic**: *epigenetic, mesogenetic, eogenetic, paragenetic, diagenetic, telogenetic, speleogenetic, ontogenetic;*

-**genic**: *epigenic, rheogenic, geogenic, basaltic, elliptic, orogenic;*

-**gean**: *hypogean, aegean, epigean/epigene, endogean;*

These adjectives determine CAUSE and the embedding methodology extracted them as candidates. However, the error originates from the seed word selection where epigenic was wrongly assigned to the LOCATION relation. Two similar examples from the same group appear in Croatian list of candidates: *epigenetski, speleogenetski, poligenetski*.

English: The adjectives like *transversal, horizontal, vertical, elliptical, sinusoidal* cannot qualify location but FORM. The two examples from the same group also appear in Croatian list of candidates: *vodoravan, konveksan.*

The adjective *sulphuric* was recognised as qualifying LOCATION while it determines COMPOSITION.

Croatian: The adjectives like *vodotopiv, vododrživ, difuzan, porozan, hidrostatski, toplinski, drenažan, tlačni, nezasićen, procjedan, direktan, ozonski, protočan* do not signify LOCATION but FUNCTION.

4.6.2 FUNCTION

In this group we found examples that were recognised in two different semantic relations FUNCTION and COMPOSITION. This entails a need for further verification of the results in order to determine whether the adjectives show multidimensional meaning or they are erroneously attributed to a certain semantic relation. Examples in English and Croatian can confirm that their isolated meaning express only COMPOSITION. Nevertheless, it is important to pay attention to the nouns they modify which can explain the meaning shift.

English: *argillaceous, carbonaceous, siliceous, dolostone, limestone, ferruginous, gaseous, hydrous, cavernous*;

Croatian: *sitnozrnat/sitnozrnast, zrnat/zrnast, krupnozrnat/ krupnozrnast, sedimentan, glinovit karbonatan, karbonatni, hidrokarbonatan, nekarbonatan, dolomitan, dolomitičan, laporovit, gipsan*.

4.6.3. COMPOSITION

Embedding results for the semantic relation COMPOSITION also include adjectives that appear as candidates in two different semantic relations but only one is correct. For example, adjectives in Croatian like *vodonepropustan, vodopropusan, neotopiv, trošiv, polupropusni, topljiv, vododržljiv, vododrživ* appear in groups COMPOSITION and FUNCTION but only FUNCTION is valid.

Adjective *monoklinski* appeared as candidate in COMPOSITION and CAUSE but it only qualifies CAUSE.

The same goes for adjectives derived from the geographical eras like *staromezozojski, gornjokretacejski, postkretacejski* and *mladomezozojski*. Their meaning qualifies CAUSE and not COMPOSITION.

Mistakes in this group also include adjectives denoting FORM instead of COMPOSITION: *dendritičan, angularan, amorfan, sigasti, heksagonski*.

We can conclude that embedding results are able to point to polysemic meaning by categorising the same candidates in two different semantic groups. In these cases, a further validation will determine if only one meaning is correct or it is possible to describe the meaning as multidimensional. In the next paragraph we present several examples of adjectives whose meaning can be attributed to two different semantic relations depending on their definition or the nouns they are modifying.

### 4.7 Multidimensional adjectives

For certain adjectives we found that two relations are possible and that they should be described as multidimensional:

English:

*carbonate, metacarbonate, noncarbonate* (COMPOSITION AND FUNCTION)

*evaporitic* (COMPOSITION AND FUNCTION)

Croatian:

*metamorfan, detritičan, neogenski, alogen, evaporitan:* CAUSE and FUNCTION

*kavernozan*: LOCATION and FORM

*pukotinski*: LOCATION and FUNCTION

*osulinski:* LOCATION and CAUSE

Furthermore, we list adjectives which vary in meaning depending on their context. They appear in two or more relations but their meaning can only be confirmed after analysing the nouns they are usually combined with.

**English:**

*igneous:* COMPOSITION, FUNCTION, CAUSE

*magmatic:* COMPOSITION, FUNCTION, CAUSE, LOCATION

*sediment:* COMPOSITION, CAUSE,

*subaerial:* LOCATION, CAUSE, FUNCTION

*siliciclastic:* COMPOSITION, FUNCTION, CAUSE

**Croatian:**

*međuzrnski:* CAUSE, COMPOSITION AND LOCATION but only COMPOSITION is valid.

*klastičan, sitnoklastičan, kataklastičan, vulkanoklastičan*: COMPOSITION, CAUSE, FUNCTION

The only example where embeddings did not suggest the right semantic relation are adjectives *gornjotrijaski, srednjotrijaski, trijaski, donjotrijaski* which were attributed relations COMPOSITION and FUNCTION but they refer to CAUSE.

## 5. Discussion and conclusions

We present a method for extracting semantically related adjectives using intersections of word embeddings and a detailed manual analysis of the extracted words. The results of the first stage show high variability in precision between relations, yet for three out of five target relations the method successfully extracts numerous meaningful adjectives pertaining to the target semantic relation.

The analysis performed in the second stage reveals several nuances of semantic similarity which we categorise into clusters. In most cases, members of a cluster share a surface linguistic component such as a suffix, prefix or word stem. Some suffixes indeed contain a semantic component pertaining to a specific relation (-genic -> CAUSE), and a shared word stem almost necessarily entails a similarity in meaning. In other cases, word embeddings allow us to retrieve synonyms with no surface similarity (*podmorski – submarinski*) only on the basis of their shared contexts.

While we cannot completely explain why a certain word is extracted as similar, we know that FastText embeddings combine the vector of the entire word and some of its substrings into a single vector. The cosine similarity thus entails also the similarity of the substrings, i.e. prefixes, suffixes and word stems.

In our future experiments we intend to extend the approach to longer expressions and multi-word terms, in particular structures which represent micro-events and which can potentially be modelled through embedding analogies.

**Acknowledgements**

The authors acknowledge the financial support from the Slovenian Research Agency for research core funding for the programme Knowledge Technologies (No.P2-0103) and the project TermFrame - Terminology and Knowledge Frames across Languages (No. J6-9372). This paper is also supported by European Union's Horizon 2020 research and innovation programme under grant agreement No. 825153, project EMBEDDIA (Cross-Lingual Embeddings for Less-Represented Languages in European News Media).

of the 13th Workshop on Building and Using Comparable Corpora, pp. 29–34. Language Resources and Evaluation Conference (LREC 2020).